\documentclass[sigconf]{aamas} 

\usepackage{balance} 
\usepackage{tikz}
\usepackage{pgf}
\usepackage[utf8]{inputenc}
\usepackage{pgfplots}
\usepackage{pgfplots}
\usepgfplotslibrary{patchplots,colormaps}
\pgfplotsset{compat=1.9} 
\DeclareUnicodeCharacter{2212}{−}
\usepgfplotslibrary{groupplots,dateplot}
\usetikzlibrary{patterns,shapes.arrows}

\usepackage{pgfplots}
\usetikzlibrary{arrows.meta}

\pgfplotsset{compat=newest,
    width=7cm,
    height=5.5cm,
    scale only axis=true,
    max space between ticks=25pt,
    try min ticks=5
}
\tikzset{
    semithick/.style={line width=0.8pt},
}
\usepgfplotslibrary{groupplots}
\usepgfplotslibrary{dateplot}

\usepackage[ruled,linesnumbered,lined,commentsnumbered]{algorithm2e}
\usepackage{cancel}

\renewcommand\footnotetextcopyrightpermission[1]{} 
\setcopyright{ifaamas}
\acmConference[AAMAS '21]{Proc.\@ of the 20th International Conference on Autonomous Agents and Multiagent Systems (AAMAS 2021)}{May 3--7, 2021}{London, UK}{U.~Endriss, A.~Now\'{e}, F.~Dignum, A.~Lomuscio (eds.)}
\copyrightyear{2021}
\acmYear{2021}
\acmDOI{}
\acmPrice{}
\acmISBN{}

\acmSubmissionID{87}

\begin{document}
\pagestyle{plain}
\title{Risk-Aware and Multi-Objective Decision Making with Distributional Monte Carlo Tree Search} 
\titlenote{This paper extends our AAMAS 2021 extended abstract \cite{hayes2021dmcts} with additional experimental results and theoretical analysis.}


\author{Conor F. Hayes}
\affiliation{%
  \institution{
       National~University~of~Ireland~Galway\\ 
        Ireland}
}
\email{c.hayes13@nuigalway.ie}

\author{Mathieu Reymond}
\affiliation{
  \institution{
        Vrije Universiteit Brussel\\
        Belgium}
}
\email{mathieu.reymond@vub.be}

\author{Diederik M.\ Roijers}
\affiliation{
  \institution{
  Vrije Universiteit Brussel (BE) \& \\ 
        HU~Univ.~of~Appl.~Sci.~Utrecht (NL)}
}
\email{diederik.yamamoto-roijers@hu.nl}

\author{Enda Howley}
\affiliation{%
  \institution{
       National~University~of~Ireland~Galway\\ 
        Ireland}
}
\email{enda.howley@nuigalway.ie}

\author{Patrick Mannion}
\affiliation{%
\institution{
       National~University~of~Ireland~Galway\\ 
        Ireland}
}
\email{patrick.mannion@nuigalway.ie}


\begin{abstract}  
In many risk-aware and multi-objective reinforcement learning settings, the utility of the user is derived from the single execution of a policy. In these settings, making decisions based on the average future returns is not suitable. For example, in a medical setting a patient may only have one opportunity to treat their illness. When making a decision, just the expected return -- known in reinforcement learning as the value -- cannot account for the potential range of adverse or positive outcomes a decision may have. Our key insight is that we should use the distribution over expected future returns differently to represent the critical information that the agent requires at decision time. 
In this paper, we propose Distributional Monte Carlo Tree Search, an algorithm that learns a posterior distribution over the utility of the different possible returns attainable from individual policy executions, resulting in good policies for both risk-aware and multi-objective settings. Moreover, our algorithm outperforms the state-of-the-art in multi-objective reinforcement learning for the expected utility of the returns. 
\end{abstract}

\keywords{Multi-objective; risk-aware; decision making; distributional; reinforcement learning; Monte Carlo tree search}  

\maketitle


\section{Introduction}
 In reinforcement learning (RL) settings, the expected return is used to make decisions. In many scenarios, the utility of a user is derived from the single execution of a policy \cite{roijers2013survey}. For example, in a medical setting a patient may only have one opportunity to select a treatment. To learn an optimal policy in these scenarios it is important to optimise under the utility of the returns. Therefore, if a policy will only be executed once, making decisions using the expected utility of the returns is not sufficient. For example, imagine we have two choices: bet or don't bet. If we bet there is a 0.5 chance of winning, which gives us a reward of 40, and a 0.5 chance of losing, which returns a reward of -20. If we do not bet, we get a reward of 10. Both of these choices have the same expected reward, with betting potentially returning a negative reward. However, once we consider a human decision maker this story becomes different; if a person receives -20 they would be in debt, and this could have a severe adverse effect on this person's well-being. While getting 40 might be nice, it may not worth the risk of going into debt. Therefore, the decision maker would prefer the non-betting strategy. As such, in order for an agent to have sufficient critical information at decision time, it is crucial to replace the expected return with a posterior distribution over the expected utility of returns. This realisation is key to risk-aware systems and, as we argue, for many multi-objective decision problems as well.

In the aforementioned scenarios, the agent must calculate the returns of a full execution of a policy before deriving the user's utility. To calculate the utility we apply the utility function to the returns where a user's utility function is known a priori. In other words, in the taxonomy of multi-objective sequential decision making by \cite{roijers2013survey}, we are in the known-weights non-linear utility setting. When optimising under the expected utility, it is critical to only apply the utility function to the returns of a full execution of a policy \cite{roijers2018multi}. This is a critical step since non-linear utility functions do not distribute across the sum of immediate and future returns. In this case, the agent must know the returns it has already accrued and the future returns before applying the utility function. For example, before the $2008$ financial crash, investment bankers were guaranteed their base salaries regardless of their losses, but their bonuses were dependent on their returns from investments. In the case of an investor incurring a loss, the only policy that would result in a bonus would be one that executes an increasingly risky strategy to win back the losses and receive some bonus.

Learning the utility of the returns is thus naturally risk-aware. Optimising the utility of the sum of the accrued and future returns to make decisions enables an agent to avoid certain undesirable outcomes. Without knowing the accrued returns, an agent cannot understand how future actions could affect the cumulative return. 
To make optimal decisions to maximise the user's utility, the agent must have information about both the accrued and the future returns.
 
A further complicating factor is that, in the real world, decision-making often involves trade-offs based on multiple conflicting objectives. For example, we may want to maximise the power output of coal-burning electrical generators while minimising $CO_{2}$ emissions. Many approaches to multi-objective decision-making only consider linear utility functions; this limitation severely restricts the real-world applicability of these methods, given that utility in many real-world problems is derived in a non-linear manner.
 
In the multi-objective case, optimising under the expected utility is known as the expected scalarised returns (ESR). For MORL, the utility function expresses the user's preferences over objectives. If the utility function is linear and is known a priori, it is possible to translate a multi-objective decision problem to its single-objective equivalent. Once translated, we can then apply single objective methods to solve the decision problem. However, if the utility function is non-linear, as human preferences often are, strictly multi-objective methods are required to find optimal solutions. We note that the ESR criterion is an understudied challenge in the MORL literature and very few methods that consider the utility of the expected return exist in the current literature. 

We propose a novel algorithm, Distributional Monte Carlo Tree Search (DMCTS), which learns a posterior distribution over the expected utility of the returns. 
DMCTS learns a posterior distribution over the utility of the returns by executing multiple individual policies and calculating the utility of the returns obtained from each policy execution. DMCTS builds upon Monte Carlo Tree Search (MCTS). 
MCTS is a heuristic search algorithm and has become a highly popular framework \cite{silver2016, Sebag2012, Kocsis2006}. Learning a posterior distribution over the utility of returns overcomes the issues present when making decisions solely with the expected return. Our key insight is that learning a posterior distribution over the utility of the returns is essential when optimising for risk-aware RL and under the MORL ESR criterion. A distribution contains more information about the range of potential negative and positive outcomes at decision time. An added feature of DMCTS is the ability to also optimise under the scalarised expected returns (SER) criterion. We implement and demonstrate DMCTS for both risk-aware and multi-objective problems. DMCTS learns good polices in risk-aware settings. Moreover, DMCTS outperforms the state-of-the-art in MORL under ESR. 

\section{Background}
\subsection{Multi-Objective Reinforcement Learning}
In multi-objective reinforcement learning (MORL), we deal with decision problems with multiple objectives, often modelled as a multi-objective Markov decision process (MOMDP). An MOMDP represents a tuple, $\mathcal{M} = (\mathcal{S}, \mathcal{A}, \mathcal{T}, \gamma, \mathcal{R})$, where $\mathcal{S}$ and $\mathcal{A}$ are the state and action spaces, $\mathcal{T} \colon \mathcal{S} \times \mathcal{A} \times \mathcal{S} \to \left[ 0, 1 \right]$ is a probabilistic transition function, $\gamma$ is a discount factor determining the importance of future rewards and $\mathcal{R} \colon \mathcal{S} \times \mathcal{A} \times \mathcal{S} \to \mathbb{R}^n$ is an $n$-dimensional vector-valued immediate reward function. In multi-objective reinforcement learning (MORL), $n>1$.

In MORL, the user's utility derives from the vector-valued outcomes (returns). This is typically modelled as a utility function that needs to be applied to these outcomes in one way or another. For this, there are two choices \cite{roijers2013survey}. Calculating the expected value of the return of a policy before applying the utility function leads to the scalarised expected returns (SER) optimisation criterion: 
\begin{equation}
    V_{u}^{\pi} = u\left(\mathbb{E} \left[ \sum\limits^\infty_{t=0} \gamma^t {\bf r}_t \:|\: \pi, \mu_0 \right]\right).
    \label{eqn:ser}
\end{equation}
SER is the most commonly used criterion in the multi-objective (single agent) planning and reinforcement learning literature \cite{roijers2013survey}. For SER, a coverage set is defined as a set of optimal solutions for all possible utility functions. If the utility function is instead applied before computing the expectation, this leads to the expected scalarised returns (ESR) optimisation criterion \cite{roijers2018multi}: 
\begin{equation}
    V_{u}^{\pi} = \mathbb{E} \left[ u\left( \sum\limits^\infty_{t=0} \gamma^t {\bf r}_t \right) \:|\: \pi, \mu_0 \right].
    \label{eqn:esr}
\end{equation}
ESR is the most commonly used criterion in the game theory literature on multi-objective games~\cite{radulescu2020survey}.

\subsection{Monte Carlo Tree Search}
One way of approaching a decision problem (in RL) is to use tree search. Perhaps the most popular of such methods is
Monte Carlo Tree Search (MCTS) \cite{Coulom2006}, which employs heuristic exploration to construct its search tree. MCTS builds a search tree of nodes, where each node has a number of children. Each child node corresponds to an action available to the agent. MCTS has two phases: the learning phase and the execution phase.

In the learning phase the agent implements the following four steps \cite{Browne2012}: selection, expansion, simulation and backpropagation. \textbf{Selection:} the agent traverses the search tree until it reaches a node that not been explored, also called a leaf node. \textbf{Expansion:} at a leaf node the node must be expanded. The agent creates a random child node and then must simulate the environment for the newly created child node. \textbf{Simulation:} the agent executes a random policy through Monte Carlo simulations until a terminal state of the environment is reached. The agent then receives the returns. \textbf{Backpropagation:} the agent must backpropagate the returns received at a terminal state to each node visited during selection where a predefined algorithm statistic e.g. UCT \cite{Coulom2006, Kocsis2006} is updated. Each step is repeated a specified number of times, which incrementally builds the search tree. Or, as we will discuss in the next subsection, a posterior belief on the returns, from which we can draw actions using Thompson sampling \cite{Bai2014}.

During the execution phase the agent must select a child node to traverse to next. The agent evaluates the statistic at each node and moves to the node which returns the maximum value; the selection phase is then re-executed. An episode ends when the execution phase arrives at a terminal state.

\label{sec:MCTS}

\subsection{(Bootstrap) Thompson Sampling}
\label{sec:BTS}
As previously mentioned, during the learning phase of MCTS, we can use Thompson sampling to take exploring actions \cite{Bai2014}. However, it is not always possible to get an exact posterior. 
In this case a bootstrap distribution over means can be used to approximate a posterior distribution \cite{efron2012,newton1995}. Eckles et al.\ \cite{EcklesKaptein2014,EckelsKaptein2019} use a bootstrap distribution to replace the posterior distribution used in Thompson Sampling. This method is known as Bootstrap Thompson Sampling (BTS) \cite{EcklesKaptein2014} and was proposed in the multi-arm bandit setting. The bootstrap distribution contains a number of bootstrap replicates, $j \in  \{1, ..., J\}$, where $J$ is a hyper-parameter that can be tuned for exploration. For a small $J$, BTS can become greedy. A larger $J$ value increases exploration,  but at a computational cost \cite{EcklesKaptein2014}. 

Each bootstrap replicate, $j$, in the bootstrap distribution contains two parameters, $\alpha_{j}$ and $\beta_{j}$, where $\frac{\alpha_{j}}{\beta_{j}}$ is an observation. At decision time, to determine the optimal action the bootstrap distribution for each arm, $i$, is sampled. The observation for the corresponding bootstrap replicate, $j$, is retrieved and the arm with the maximum observation is pulled \cite{EcklesKaptein2014}.  

The distribution which corresponds to the maximum arm is randomly re-weighted by simulating a coin-flip (commonly known as sampling from a Bernoulli bandit) for each bootstrap replicate, $j$, in the bootstrap distribution. If the coin-flip is heads, the observation for $j$ is re-weighted. To re-weight an observation, the return is added to the $\alpha_{j}$ value and $1$ is added to $\beta_{j}$ \cite{EcklesKaptein2014}. 

Bootstrap methods with random re-weighting \cite{Rubin1981} are more computationally appealing as they can be conducted online rather than re-sampling data \cite{Oza2005}. BTS addresses problems of scalability and robustness when compared to Thompson Sampling \cite{EcklesKaptein2014}.

\subsection{Expected Utility Policy Gradient}
Expected Utility Policy Gradient (EUPG) is a MORL algorithm for ESR \cite{roijers2018multi}. EUPG is an extension of Policy Gradient \cite{Sutton1999, williams1992}, where Monte Carlo simulations are used to compute the returns and optimise the policy. EUPG calculates the accrued returns, $\textbf{R}^-_{t}$, which is the sum of the immediate returns received as far as the current timestep, $t$. EUPG also calculates the future returns, $\textbf{R}^+_{t}$, which is the sum of the immediate returns from the current timestep, $t$, to the terminal state. Using both the accrued and future returns enables EUPG to optimise over the utility of the full returns of an episode, where utility function is applied the sum of $\textbf{R}^-_{t}$ and $\textbf{R}^+_{t}$. 

Roijers et al.\ \cite{roijers2018multi} showed for ESR the accrued and future returns must be considered when learning. Applying this consideration to EUPG, the algorithm achieves the state-of-the-art performance under ESR. In this paper, we use the same method of adding past and future returns together before applying the utility inside of the MCTS search scheme.

\section{Distributional Monte Carlo Tree Search}
\label{sec:DMCTS}
The majority of RL research focuses on learning an optimal policy based on the expected returns, known as the value. Under the expected scalarised returns (ESR), a single execution of a policy is used to derive the utility of a user. In the outlined scenario, it is crucial that the agent has sufficient information at decision time to exploit positive outcomes and avoid negative outcomes. Under ESR, taking actions based on the expected returns fails to provide the agent with this information. However, acting based on a distribution over the expected utility of returns overcomes this issue. In Section \ref{sec:DMCTS}, we present our Distributional Monte Carlo Tree Search (DMCTS) algorithm which learns a posterior distribution over the expected returns.

DMCTS builds an expectimax search tree through the same process as MCTS. A search tree is a representation of the state-action space that is incrementally built though Monte Carlo simulations. The search tree is built using nodes, where a node represents an action available to the agent. Each node has a number of children corresponding to the number of actions available (see Section \ref{sec:MCTS}). 

An expectimax search tree \cite{Veness2011} uses both decision and chance nodes. Each decision node represents a state, action and reward of an MOMDP, and has a child chance node per action. In this paper we examine only environments with stochastic rewards. Each chance node represents the state and action of an MOMDP. At each chance node, the environment is sampled. If an unseen reward is generated when sampling the environment, a new child decision node is created. This process repeats as the agent traverses the search tree. It is important to note that each chance node and its parent decision node share the same state and action. A child decision node is only created when an unseen reward is observed from sampling the environment. DMCTS uses the phases of selection, expansion, simulation and backpropagation to build and traverse the search tree, similar to MCTS (Section \ref{sec:MCTS}).

Usually an expectation of the returns is maintained at each chance node, and the agent seeks to maximise the expectation. Making decisions based on the expected returns does not account for potential undesired outcomes. For risk-aware RL and MORL under ESR, we need to be able to make decisions with sufficient information to avoid such undesirable outcomes. Under these conditions, an alternative to making decisions based on the expected returns must be found.

Learning a posterior distribution over the utility of the returns can be used to replace the expected future returns (of vanilla MCTS) at each node. We outline our algorithm for single-objective risk-aware RL and MORL under ESR. With minor changes to our algorithm we can also apply DMCTS to multi-objective RL under SER, which we will discuss briefly at the end of this section.

To compute the distribution we first calculate the accrued returns, $\textbf{R}^-_{t}$. The accrued returns is the sum of returns received during the execution phase as far as timestep, $t$, where $\textbf{r}_t$ is the reward received at each timestep,
\begin{equation*}
    \textbf{R}^-_{t} = \sum_{0}^{t-1} \textbf{r}_{t}.
\end{equation*}

Secondly, we must calculate future returns, $\textbf{R}^+_{t}$. The future returns is the sum of the rewards received when traversing the search tree during the learning phase and Monte Carlo simulations from timestep, $t$, to a terminal node, $t_{n}$, 
\begin{equation*}
    \textbf{R}^+_{t} = \sum_{t}^{t_{n}} \textbf{r}_{t}.
\end{equation*}
The cumulative returns, $\textbf{R}_t$, is the sum of the accrued returns, $\textbf{R}^-_{t}$, and the expected future returns, $\textbf{R}^+_{t}$,
\begin{equation*}
    \textbf{R}_{t} = \textbf{R}^-_{t} + \textbf{R}^+_{t}.
\end{equation*}
$\textbf{R}_{t}$ is backpropagated to each node in the search tree, where the utility is computed, $u(\textbf{R}_{t})$. Since non-linear utility functions do not distribute across the sums for the immediate or future returns, we must calculate the cumulative returns, $\textbf{R}_{t}$. Applying the utility function to the cumulative returns, $\textbf{R}_{t}$, ensures we satisfy the ESR criterion.
In single-objective RL where risk is not considered, the expected future returns are sufficient to base the optimal action on. By contrast, for risk-aware and ESR-MORL it is essential to use the cumulative returns ${\bf R}_t$ to determine the optimal action, as we have argued before. In this paper, we do not use discounting as we perform evaluations only on finite horizon tasks. We note that DMCTS can easily be adapted to discounted settings.

At each node we aim to maintain a posterior distribution over the expected utility of the returns. However, because the utility function may be non-linear, a parametric form of the posterior distribution may not exist. Since a bootstrap distribution can be used to approximate a posterior \cite{efron2012,newton1995}, it is much more suitable to maintain a bootstrap distribution over the expected utility of the returns at each node. 

Each bootstrap distribution contains a number of bootstrap replicates, $j \in  \{1, ..., J\}$ \cite{EcklesKaptein2014} (See Section \ref{sec:BTS}). On initialisation of a new node, for each bootstrap replicate, $j$, the parameters $\alpha_{j}$ and $\beta_{j}$ are both set to $1$. Moreover, $\alpha_{j}$ can be set to positive values to increase initial exploration without a computational cost. 

During the backpropagation phase the bootstrap distribution at each node is updated. Algorithm \ref{alg:DMCTS_UpdateDistribution} outlines how a bootstrap distribution for a node is updated. At node $i$, for each bootstrap replicate, $j$, a coin flip is simulated (See Algorithm \ref{alg:DMCTS_UpdateDistribution}, Line \ref{alg:UpdateDistribution:line:bernoulli}). If the result of the coin flip is equal to $1$ (heads), $\alpha_{ij}$ and $\beta_{ij}$ are updated:
\begin{equation*}
    \alpha_{ij} = \alpha_{ij} +  u(\textbf{R}_t)
\end{equation*}
\begin{equation*}
    \beta_{ij} = \beta_{ij} + 1
\end{equation*}

To select actions while learning, we use the previously computed statistics. At each timestep the agent must choose which action to execute in order to traverse the search tree (as outlined in Algorithm \ref{alg:DMCTS_ThompsonSample}). At node $n$, we select an action by sampling the bootstrap distribution at each child node, $i$. For each sampled bootstrap replicate, $j$, the $\alpha_{ij}$ and $\beta_{ij}$ values are retrieved and $\frac{\alpha_{ij}}{\beta_{ij}}$ is computed. 
Since the following is true,
\begin{equation}
    \frac{\alpha_{ij}}{\beta_{ij}} \equiv \mathbb{E}[u(\textbf{R}^-_{t} + \textbf{R}^+_{t})],
\label{eqn:alphabeta_ESR}
\end{equation}
by maximising over $i$ in Equation \ref{eqn:alphabeta_ESR}, we select an action corresponding to $j$ approximately proportionally to the probability of that action being optimal -- as per the Bootstrap Thompson Sampling exploration strategy. 
The agent then executes the action, $a^*$, which corresponds to the following:
\begin{equation*}
 a^* = \arg \max_{i} \frac{\alpha_{ij}}{\beta_{ij}}.
\end{equation*}
We note that at execution time we can simply select the overall maximising action by averaging over all the acquired data (ignoring the bootstrap replicates), thereby maximising the ESR criterion:
\begin{equation}
    ESR = \mathbb{E}[u(\textbf{R}^-_{t} + \textbf{R}^+_{t})].
\label{eqn:ESR_DMCTS}
\end{equation}

Using the outlined algorithm, DMCTS is able to learn optimal policies for risk-aware settings and under ESR for multi-objective settings. In Section \ref{sec:Experiments} we have evaluated DMCTS for risk-aware settings and multi-objective settings under ESR.

The majority of MORL research focuses on the SER criterion rather than the ESR criterion \cite{roijers2013survey}. With a minor change to the algorithm it is also possible for DMCTS to optimise for the SER criterion. Specifically, under the SER criterion we maintain a bootstrap distribution over expected return vectors. For a node under SER, it is important to ensure that $\boldsymbol\alpha$ is initialised to a vector for each bootstrap replicate, $j$. The number of values in the bootstrap replicate vector, $\boldsymbol\alpha_{j}$, corresponds to the number of objectives, $o$, where each value is set to $1$, $\boldsymbol\alpha_{j} = [1, ..., 1_{o}]$. The parameter $\beta$ is set to 1 for each bootstrap replicate, $j$.

To update the bootstrap distribution of node, $i$, we use the same process as under ESR (see Algorithm \ref{alg:DMCTS_UpdateDistribution}). At each node a coin flip is simulated for each bootstrap replicate, $j$. If the simulated coin flip returns $1$ (heads), then we update the bootstrap replicate. We use the following to update $\boldsymbol\alpha_{ij}$ and $\beta_{ij}$:
\begin{equation*}
    \boldsymbol\alpha_{ij} = \boldsymbol\alpha_{ij} + \textbf{R}_t,
    \label{eqn:ser_alpha_update}
\end{equation*}
\begin{equation*}
    \beta_{ij} = \beta_{ij} + 1.
    \label{eqn:ser_beta_update}
\end{equation*}

At learning time, we sample the bootstrap distribution at each child node, $i$. For a sampled bootstrapped replicate, $j$, the parameters $\boldsymbol{\alpha}_{ij}$ and $\beta_{ij}$ are retrieved. Before we can determine the optimal action, we must compute $\frac{\boldsymbol\alpha_{ij}}{\beta_{ij}}$.
We then apply the utility function, $u$, to $\frac{\boldsymbol\alpha_{ij}}{\beta_{ij}}$ to compute the utility of the expected returns. Since,
\begin{equation}
    u(\frac{\boldsymbol\alpha_{ij}}{\beta_{ij}}) \equiv u(\mathbb{E}[\textbf{R}^-_{t} + \textbf{R}^+_{t}]),
\label{eqn:alphabeta_rt}
\end{equation}
the agent can then execute the action, $a^*$, which corresponds to the following:
\begin{equation*}
 a^* = \arg \max_{i} u(\frac{\boldsymbol\alpha_{ij}}{\beta_{ij}}).
\end{equation*}
\begin{algorithm}
\SetAlgoLined
\textbf{Input}: i\ $\leftarrow$ Node\ in\ the\ tree\ \\
\textbf{Input}: $R_{t}\ \leftarrow$ Cumulative\ Reward\ \\
J\ $\leftarrow$ node.bootstrapDistribution \\
\For{j,\ ...,\ $J$\ bootstrap\ replicates}{
Sample\ $d_{j}$\ from\ Bernoulli(1/2) \label{alg:UpdateDistribution:line:bernoulli} \\
\If{$d_{j}$ = 1}{
$\alpha_{ij} = \alpha_{ij} + R_{t}$ \\
$\beta_{ij} = \beta_{ij} + 1$ \\
}}
\caption{UpdateDistribution}
\label{alg:DMCTS_UpdateDistribution}
\end{algorithm}
\begin{algorithm}
\SetAlgoLined
\textbf{Input}: n\ $\leftarrow$ Node\ in\ the\ tree\ \\
\textbf{Require}: $\alpha$,\ $\beta$\ prior\ paramaeters\ \\
$\alpha_{ij}$ := $\alpha$,\ $\beta_{ij}$ := $\beta$\  \{For\ each\ n\ child,\ $i$,\ and\ each\ bootstrap\ replicate,\ $j$ \} \\
\For{$i$,\ ...,\ n\ children} 
{
Sample\ $j$\ from\ uniform\ 1,\ ...,\ $J$\ bootstrap\ replicates\\
Retrieve\ $\alpha_{ij},\ \beta_{ij}$ \\
}
maxChild\ = $\arg \max_{i} \frac{\alpha_{ij}}{\beta_{ij}} $\\
\textbf{return} maxChild\ or\ maxChild.action
\caption{ThompsonSample}
\label{alg:DMCTS_ThompsonSample}
\end{algorithm}
\section{Experiments}
In order to evaluate our DMCTS algorithm, we test DMCTS in multiple settings. Firstly, we evaluate DMCTS in a risk-aware setting. Secondly, we evaluate DMCTS in multi-objective settings under both ESR and SER. In multi-objective settings we test our algorithm on variants of standard benchmark problems from the MORL literature.
At each timestep for DMCTS, the learning phase is performed multiple times before an action is selected during the execution phase. To fairly evaluate all other algorithms against DMCTS, we have altered each benchmark algorithm to have the same number of policy executions of each environment at each timestep as DMCTS. So at each timestep, each algorithm gets $n_{exec}$ full policy executions worth of learning from that state and timestep onward. For the other algorithms (except DMCTS) this has the effect of increasing the learning speed. The number of policy executions $n_{exec}$ varies for each problem domain. 
All experiments are averaged over 10 runs. 
\label{sec:Experiments}
\subsection{Risk-Aware MDP}
Before testing DMCTS on benchmark problems from the MORL literature, we evaluate DMCTS in a risk-aware problem domain under ESR. Shen et al.\ \cite{Yun2014} define a risk-aware MDP where an agent must decide from a number of stocks in which to invest. The underlying MDP which has $4$ actions (each action is a monetary amount, in Euros, of investment) and $7$ states. At each timestep the agent must select a monetary amount to invest in the stock for a given state. We can invest \texteuro 0, \texteuro 1, \texteuro 2 or \texteuro 3 in a stock at each timestep. Each stock has a probability of making a profit and a probability of making a loss where the agent's return is the action multiplied by the stock price. In the risk-aware MDP, certain policies are risk-averse, risk-seeking or a mixture of both. For example, if an agent takes action 0 or action 1 at each timestep, the agent is said to be risk-averse. Executing action 0 at each timestep is the most risk-averse policy that an agent can learn. Investing \texteuro 0 at each timestep means the agent has no gains or losses, given the returns are equal to the monetary investment multiplied by the stock price. The type of policy an agent learns depends on the utility function. For certain utility functions, the agent could be risk-seeking or risk-averse. To evaluate our DMCTS algorithm we use the following risk-averse non-linear utility function:
\begin{equation}
    u = 1 - e^{-r_{t}}.
\end{equation}
In the risk-aware setting, we compare our algorithm against Q-learning, where we aim to learn the policy that is risk-averse. 
The parameter $n_{exec}$ is set to $10$ for each algorithm and each experiment lasts for $1,000$ episodes.
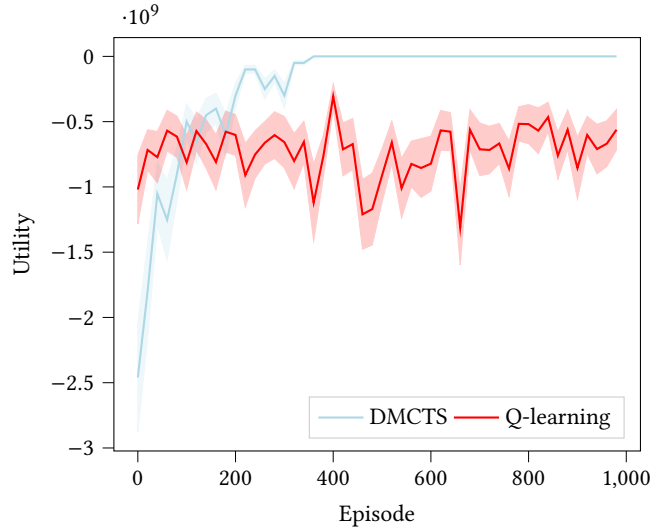
\begin{figure}[t]
    \centering
\begin{tikzpicture}

\definecolor{color0}{rgb}{0.67843137254902,0.847058823529412,0.901960784313726}

\begin{axis}[
legend cell align={left},
legend columns=3,
legend style={fill opacity=0.8, draw opacity=1, text opacity=1, at={(0.97,0.03)}, anchor=south east, draw=white!80!black},
tick align=outside,
tick pos=left,
x grid style={white!69.0196078431373!black},
xlabel={Episode},
xmin=-49, xmax=1029,
xtick style={color=black},
y grid style={white!69.0196078431373!black},
ylabel={Utility},
ymin=-3023033929.755, ymax=143953996.655,
ytick style={color=black}
]
\path [draw=color0, fill=color0, opacity=0.2]
(axis cs:0,-2879079933.1)
--(axis cs:0,-2040920066.9)
--(axis cs:20,-1435128171.6)
--(axis cs:40,-810603690.8)
--(axis cs:60,-947372014.3)
--(axis cs:80,-690840201.4)
--(axis cs:100,-362967967.8)
--(axis cs:120,-479933460.2)
--(axis cs:140,-318238430.6)
--(axis cs:160,-284049818.9)
--(axis cs:180,-433837784.2)
--(axis cs:200,-205142846.3)
--(axis cs:220,-68377223.29)
--(axis cs:240,-68377223.29)
--(axis cs:260,-170943058.38)
--(axis cs:280,-102565834.98)
--(axis cs:300,-205131670.08)
--(axis cs:320,-34188611.66)
--(axis cs:340,-34188611.7)
--(axis cs:360,0)
--(axis cs:380,0)
--(axis cs:400,0)
--(axis cs:420,0)
--(axis cs:440,0)
--(axis cs:460,0)
--(axis cs:480,0)
--(axis cs:500,0)
--(axis cs:520,0)
--(axis cs:540,0)
--(axis cs:560,0)
--(axis cs:580,0)
--(axis cs:600,0)
--(axis cs:620,0)
--(axis cs:640,0)
--(axis cs:660,0)
--(axis cs:680,0)
--(axis cs:700,0)
--(axis cs:720,0)
--(axis cs:740,0)
--(axis cs:760,0)
--(axis cs:780,0)
--(axis cs:800,0)
--(axis cs:820,0)
--(axis cs:840,0)
--(axis cs:860,0)
--(axis cs:880,0)
--(axis cs:900,0)
--(axis cs:920,0)
--(axis cs:940,0)
--(axis cs:960,0)
--(axis cs:980,0)
--(axis cs:980,0)
--(axis cs:980,0)
--(axis cs:960,0)
--(axis cs:940,0)
--(axis cs:920,0)
--(axis cs:900,0)
--(axis cs:880,0)
--(axis cs:860,0)
--(axis cs:840,0)
--(axis cs:820,0)
--(axis cs:800,0)
--(axis cs:780,0)
--(axis cs:760,0)
--(axis cs:740,0)
--(axis cs:720,0)
--(axis cs:700,0)
--(axis cs:680,0)
--(axis cs:660,0)
--(axis cs:640,0)
--(axis cs:620,0)
--(axis cs:600,0)
--(axis cs:580,0)
--(axis cs:560,0)
--(axis cs:540,0)
--(axis cs:520,0)
--(axis cs:500,0)
--(axis cs:480,0)
--(axis cs:460,0)
--(axis cs:440,0)
--(axis cs:420,0)
--(axis cs:400,0)
--(axis cs:380,0)
--(axis cs:360,0)
--(axis cs:340,-65811388.3)
--(axis cs:320,-65811388.28)
--(axis cs:300,-394868329.72)
--(axis cs:280,-197434164.82)
--(axis cs:260,-329056941.42)
--(axis cs:240,-131622776.55)
--(axis cs:220,-131622776.55)
--(axis cs:200,-394889843.5)
--(axis cs:180,-771013867.4)
--(axis cs:160,-515950180.7)
--(axis cs:140,-581761569)
--(axis cs:120,-824918191.4)
--(axis cs:100,-637032031.8)
--(axis cs:80,-1114011449.8)
--(axis cs:60,-1557545017.7)
--(axis cs:40,-1294379163.2)
--(axis cs:20,-2164871828.4)
--(axis cs:0,-2879079933.1)
--cycle;

\path [draw=red, fill=red, opacity=0.2]
(axis cs:0,-1282812680)
--(axis cs:0,-757187320)
--(axis cs:20,-562423693)
--(axis cs:40,-575139619.9)
--(axis cs:60,-413354426.3)
--(axis cs:80,-457633746.8)
--(axis cs:100,-602560439.3)
--(axis cs:120,-425849617.2)
--(axis cs:140,-485634393.2)
--(axis cs:160,-598653819.2)
--(axis cs:180,-419755321.5)
--(axis cs:200,-446147714.8)
--(axis cs:220,-666195094.9)
--(axis cs:240,-569801027.4)
--(axis cs:260,-499525685.3)
--(axis cs:280,-423909778.5)
--(axis cs:300,-460515622.5)
--(axis cs:320,-590409179.9)
--(axis cs:340,-496697456.4)
--(axis cs:360,-825974992.4)
--(axis cs:380,-601965958.6)
--(axis cs:400,-210664748.5)
--(axis cs:420,-509219247)
--(axis cs:440,-475607873.2)
--(axis cs:460,-942275917.5)
--(axis cs:480,-894837453.7)
--(axis cs:500,-706620361.1)
--(axis cs:520,-495266020.8)
--(axis cs:540,-781327426.2)
--(axis cs:560,-648835292.2)
--(axis cs:580,-629865102.1)
--(axis cs:600,-611571658.4)
--(axis cs:620,-414004275.5)
--(axis cs:640,-431039258.8)
--(axis cs:660,-1022375399.3)
--(axis cs:680,-408744759.1)
--(axis cs:700,-510392544.2)
--(axis cs:720,-535022115.1)
--(axis cs:740,-513620726.5)
--(axis cs:760,-648308357)
--(axis cs:780,-385967820.3)
--(axis cs:800,-369138211.7)
--(axis cs:820,-393620386.6)
--(axis cs:840,-351944148.7)
--(axis cs:860,-579453062)
--(axis cs:880,-410011150.5)
--(axis cs:900,-617166461.3)
--(axis cs:920,-459457585.9)
--(axis cs:940,-519739916.9)
--(axis cs:960,-493619034.3)
--(axis cs:980,-406979010.4)
--(axis cs:980,-717020989.6)
--(axis cs:980,-717020989.6)
--(axis cs:960,-844380965.7)
--(axis cs:940,-900260083.1)
--(axis cs:920,-744542414.1)
--(axis cs:900,-1092833538.7)
--(axis cs:880,-713988849.5)
--(axis cs:860,-940546938)
--(axis cs:840,-578055851.3)
--(axis cs:820,-744379613.4)
--(axis cs:800,-668861788.3)
--(axis cs:780,-648032179.7)
--(axis cs:760,-1071691643)
--(axis cs:740,-820379273.5)
--(axis cs:720,-898977884.9)
--(axis cs:700,-913607455.8)
--(axis cs:680,-715255240.9)
--(axis cs:660,-1597624600.7)
--(axis cs:640,-722960741.2)
--(axis cs:620,-719995724.5)
--(axis cs:600,-1032428341.6)
--(axis cs:580,-1080134897.9)
--(axis cs:560,-999164707.8)
--(axis cs:540,-1238672573.8)
--(axis cs:520,-824733979.2)
--(axis cs:500,-1117379638.9)
--(axis cs:480,-1445162546.3)
--(axis cs:460,-1477724082.5)
--(axis cs:440,-868392126.8)
--(axis cs:420,-914780753)
--(axis cs:400,-403335251.5)
--(axis cs:380,-922034041.4)
--(axis cs:360,-1414025007.6)
--(axis cs:340,-813302543.6)
--(axis cs:320,-1013590820.1)
--(axis cs:300,-853484377.5)
--(axis cs:280,-782090221.5)
--(axis cs:260,-824474314.7)
--(axis cs:240,-934198972.6)
--(axis cs:220,-1153804905.1)
--(axis cs:200,-757852285.2)
--(axis cs:180,-734244678.5)
--(axis cs:160,-1021346180.8)
--(axis cs:140,-858365606.8)
--(axis cs:120,-718150382.8)
--(axis cs:100,-1021439560.7)
--(axis cs:80,-772366253.2)
--(axis cs:60,-724645573.7)
--(axis cs:40,-968860380.1)
--(axis cs:20,-871576307)
--(axis cs:0,-1282812680)
--cycle;

\addplot [semithick, color0]
table {%
0 -2460000000
20 -1800000000
40 -1052491427
60 -1252458516
80 -902425825.6
100 -499999999.8
120 -652425825.8
140 -449999999.8
160 -399999999.8
180 -602425825.8
200 -300016344.9
220 -99999999.92
240 -99999999.92
260 -249999999.9
280 -149999999.9
300 -299999999.9
320 -49999999.97
340 -50000000
360 0
380 0
400 0
420 0
440 0
460 0
480 0
500 0
520 0
540 0
560 0
580 0
600 0
620 0
640 0
660 0
680 0
700 0
720 0
740 0
760 0
780 0
800 0
820 0
840 0
860 0
880 0
900 0
920 0
940 0
960 0
980 0
};
\addlegendentry{DMCTS}
\addplot [semithick, red]
table {%
0 -1020000000
20 -717000000
40 -772000000
60 -569000000
80 -615000000
100 -812000000
120 -572000000
140 -672000000
160 -810000000
180 -577000000
200 -602000000
220 -910000000
240 -752000000
260 -662000000
280 -603000000
300 -657000000
320 -802000000
340 -655000000
360 -1120000000
380 -762000000
400 -307000000
420 -712000000
440 -672000000
460 -1210000000
480 -1170000000
500 -912000000
520 -660000000
540 -1010000000
560 -824000000
580 -855000000
600 -822000000
620 -567000000
640 -577000000
660 -1310000000
680 -562000000
700 -712000000
720 -717000000
740 -667000000
760 -860000000
780 -517000000
800 -519000000
820 -569000000
840 -465000000
860 -760000000
880 -562000000
900 -855000000
920 -602000000
940 -710000000
960 -669000000
980 -562000000
};
\addlegendentry{Q-learning}
\end{axis}

\end{tikzpicture}
    \caption{Results from the risk-aware environment. Learning a bootstrap distribution over the expected utility of the returns (DMCTS) is critical to learning the optimal risk-averse policy for a risk-averse utility function.}
    \label{fig:riskaverse_utility}
\end{figure}

As shown in Figure \ref{fig:riskaverse_utility}, DMCTS consistently learns the optimal policy for the above risk-averse utility function. The policy, which avoids all risk, has a cumulative utility of 0. DMCTS needs around $400$ episodes to converge to the optimal policy, while Q-learning struggles to learn a stable policy for the given utility function. Maintaining a bootstrap distribution over the expected utility of the returns enables DMCTS to avoid all risk. The ability for an agent to access a distribution when learning ensures the agent can make more informed decisions to maximise its utility which, in this case, is risk-averse. 

\subsection{Fishwood}
To evaluate DMCTS in a multi-objective setting under ESR, we use a number of problem domains. Firstly, we evaluate DMCTS in the Fishwood problem \cite{roijers2018multi}, given this is one of the very few domains for which ESR results have been published. In Fishwood the agent has two states: at the river or in the woods, two actions: move to the other state or stay, and two objectives: to catch fish (when at the river) and obtain wood (when in the woods). The Fishwood environment is parameterised by the probabilities of successfully obtaining fish and wood at these respective states. In this paper we use the following values: at the river the agent has a $0.25$ chance of catching a fish and in the woods the agent has a $0.65$ chance of acquiring wood. For every fish caught, two pieces of wood are required to cook the fish, which results in a utility of $1$. The goal in this setting is to maximise the following non-linear utility function:
\begin{equation}
    u = \min \left( \texttt{fish}, \left\lfloor \frac{\texttt{wood}}{2} \right\rfloor \right).
\end{equation}

To maximise utility in Fishwood it is essential that both past and future returns are taken into consideration when learning. For example, if there are $5$ timesteps remaining and the agent has received $2$ pieces of wood, the agent should go to the river and try to catch a fish to ensure a utility of $1$ \cite{roijers2018multi}.

To evaluate DMCTS in the Fishwood domain, we compare DMCTS against C51, Expected Utility Policy Gradient (EUPG) \cite{roijers2018multi}, and Q-learning. EUPG achieves state-of-the-art results in the Fishwood problem under ESR \cite{roijers2018multi}. C51 \cite{bellemare2017distributional} is a distributional deep reinforcement learning algorithm that achieved state-of-the-art results in the Atari game problem domain.

For C51 the learning parameters were set as follows: $V_{min} = 0$, $V_{max} = 2$, $\epsilon = 0.1$, $\gamma = 1$ and $\alpha = 0.1$. For Q-learning, the learning parameters were set as follows: $\epsilon = 0. 1$, $\gamma = 1$ and $\alpha = 0.1$. For DMCTS we set the $\alpha_{j}$ parameter to $10$ for each bootstrap replicate, $j$. We set $n_{exec} = 2$ and ran each experiment for $20,000$ episodes where each episode has $13$ timesteps.
\begin{figure}[t]
    \centering
    \input{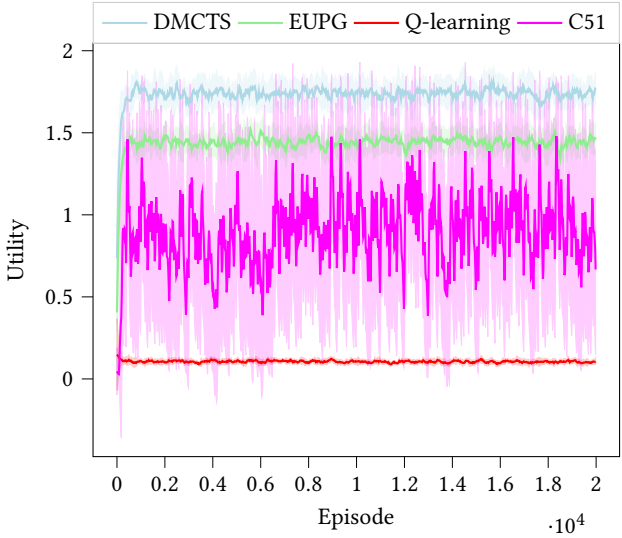}
    \caption{Results from the Fishwood environment where DMCTS achieves state-of-the-art performance in a multi-objective setting over EUPG.}
    \label{fig:fishwood_utility}
\end{figure}
As shown in Figure \ref{fig:fishwood_utility}, Q-learning and C51 fail to learn any meaningful policy. The utility for C51 fluctuates throughout experimentation and fails to learn a consistent policy, while the utility for Q-learning remains close to zero. This is because Q-learning and C51 do not take the accrued returns into consideration when learning, which has a negative impact on the ability of both algorithms to learn in the Fishwood domain.

By contrast, DMCTS and EUPG outperform both C51 and Q-learning. DMCTS achieves a higher utility when compared to EUPG. Both algorithms use Monte Carlo simulations of the environment and optimise over the expected utility of the returns of a full episode. Although both algorithms use Monte Carlo simulations of the environment, policy gradient algorithms are sample inefficient. DMCTS is sample efficient since DMCTS shares the learning phase steps with MCTS, which has been shown to be sample efficient \cite{Abramson1987, Chang2005}. In the Fishwood environment, the agent is not guaranteed to obtain a fish or a piece of wood. For an action in a particular state the agent may need multiple simulations to understand the underlying distribution of the stochastic rewards. Since DMCTS builds an expectimax tree, the agent can re-sample the environment at each chance node when learning. With repeated sampling at each chance node and Monte Carlo simulations, the agent can build an accurate bootstrap distribution over the expected utility of returns. Using the learned bootstrap distribution over the expected utility of the returns enables DMCTS to outperform EUPG and achieve state-of-the-art performance under ESR.

\subsection{Renewable Energy Dynamic Economic Emissions Dispatch}
Next, we evaluate our DMCTS algorithm in a complex problem domain with a large state action space. Renewable Energy DEED (REDEED) is a variation of the traditional DEED problem \cite{BasuDEED}. In REDEED, the power demand for a city must be met over $24$ hours. To supply the city with sufficient power, a number of generators are required. There are $9$ fossil fuel-powered generators, including a slack generator and $1$ generator powered by renewable energy which is generated by a wind turbine. The optimal power output for each generator was derived by Mannion et al.\ \cite{mannion2018reward} and the derived values are used for the both the fossil fuel generators and the renewable energy generator. In this example, Generator 3 is controlled by an agent, Generator 1 is a slack generator and Generator 4 is powered by a wind turbine.

In this setting we imagine a period of $24$ hours and for each hour we receive a weather forecast for a city. For hours $1 - 15$, the weather is predictable and the optimal power values derived by Mannion et al.\ \cite{mannion2018reward} can be used to generate power. From hours $16 - 24$, a storm is forecast for the city. During the storm, both high and low levels of wind are expected and the weather forecast impacts how much power the wind turbine can generate. At each hour during the storm, there is a 0.15 chance the wind turbine will produce $25\%$ less power than optimal, a 0.7 chance the wind turbine will produce optimal power and a 0.15 chance the wind turbine will produce $25\%$ more power than optimal. In the REDEED problem we aim to learn an optimal policy that can ensure the required power is met over the entire day while reducing both the cost and emissions created by all generators.

The goal is to maximise the following linear utility function under the ESR criterion,
\begin{equation}
    \label{eqn:linear_scalarisation}
    R_{+} = -\sum_{o=1}^{O} w_{o}f_{o},
\end{equation}
where $w_o$ is the objective weight, and $f_{o}$ is the objective function. The objective weights used are $w_{c} = 0.225$, $w_{e} = 0.275$ and $w_{p} = 0.5$ \cite{mannion2018reward}.

The following equation calculates the local cost for each generator $n$, at each hour $m$:
\begin{equation}
        f_c^L(n,m) = a_n + b_n P_{nm} + c_n (P_{nm})^2 + |d_n sin\{ e_n (P_{n}^{min} - P_{nm}) \}|. 
\end{equation}%
Therefore the global cost for all generators can be defined as:
\begin{equation}
        f_c^G(m) = \sum\limits_{n=1}^N f_c^L(n,m).
\end{equation}
The local emissions for each generator, $n$, at each hour, $m$, is calculated using the following equation :
\begin{equation}
        f_e^L(n,m) = E(a_n + b_n P_{nm} + \gamma_n(P_{nm})^2 + \eta \exp \delta P_{nm}).
\end{equation}%
Therefore the global emissions for all generators can be defined as:
\begin{equation}
        f_e^G(m) = \sum\limits_{n=1}^N f_e^L(n,m).
\end{equation}%
It is important to note the emissions for the generator controlled by the wind turbine are set to $0$.

If the agent exceeds the ramp and power limits a penalty is received. A global penalty function $f_p^G$ is defined to capture the violations of these constraints,
\begin{equation}
    f_{p}^G(m) = \sum\limits_{v=1}^V C(|h_{v} + 1| \delta_v).
    \label{eqn:penaltyHour}
\end{equation}

Along with cost and emissions, the penalty function is an additional objective that will need to be optimised. All equations and parameters absent from this paper that are required to implement this problem domain can be found in the works of Basu \cite{BasuDEED} and Mannion et al.\ \cite{mannion2018reward}.

To evaluate DMCTS in the REDEED domain, we compare DMCTS against EUPG, C51, and Q-learning \cite{mannion2018reward,HayesTLO}.

For C51 the learning parameters were set as follows: $V_{min} = -1.75e^{9}$, $V_{max} = 0$, $\epsilon = 0.1$, $\gamma = 1$ and $\alpha = 0.1$. For Q-learning, the learning parameters were set as follows: $\epsilon = 0. 1$, $\gamma = 1$ and $\alpha = 0.1$. For the REDEED problem the agent learns for $10,000$ episodes and $n_{exec} = 10$ for each algorithm.

\begin{figure}[t]
    \input{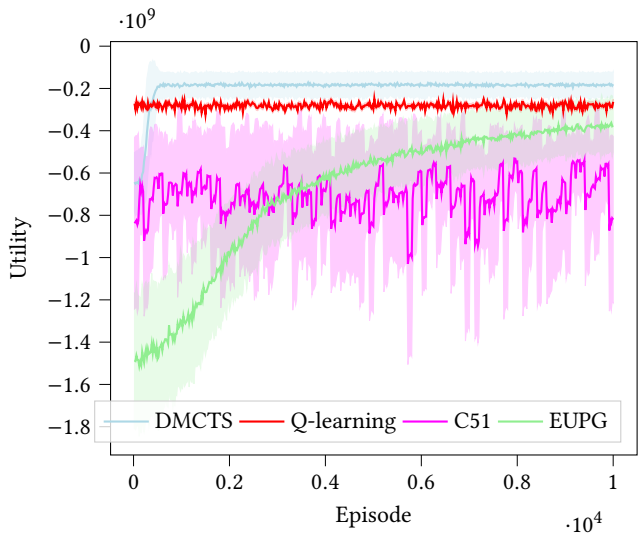}
    \caption{Results from the REDEED environment DMCTS outperforms EUPG, C51 and Q-learning. DMCTS achieves a higher utility compared to other algorithms used throughout experimentation in the REDEED domain under ESR.}
    \label{fig:deed_utility}
\end{figure}
As seen in Figure \ref{fig:deed_utility}, DMCTS outperforms EUPG, Q-learning and C51 in the REDEED domain. C51 struggles to learn a consistent policy and C51's utility fluctuates throughout experimentation. The hyper-parameters chosen for C51 provide the most optimal performance but are difficult to tune. Although the learning speed of EUPG is slow, EUPG achieves a higher utility than C51 but does not achieve a utility as high as Q-learning or DMCTS.

Although Q-learning outperforms EUPG and C51 in the REDEED environment, it does not achieve a higher utility when compared with DMCTS. C51 makes decisions based on a distribution over the expected returns and Q-learning makes decisions based on the expected future returns; due to this, both algorithms fail to learn good policies under ESR because they do not take both accrued and future returns into consideration. 

The results presented in this paper evaluate C51 in an environment with a large state action space and complex returns. We hypothesise that a reason for poor performance is C51's inability to learn a distribution over the full returns and the level of discretisation of the distribution. The distribution for C51 uses 51 bins to discretise the algorithm's distribution. Bellemare et al.\ \cite{bellemare2017distributional} claim this parameter for discretisation is optimal. However, the results presented in this paper show this parameter setting is not optimal in scenarios where the returns are not simple scalars over small ranges. How C51 would perform using different numbers of bins other than the 51 recommended by Bellemare et al.\ \cite{bellemare2017distributional} is an open question, which we do not address here as it is outside the scope of this work. The results present in Figure \ref{fig:deed_utility} show that C51 struggles to scale to large problem domains with complex returns over large ranges. 
Instead, DMCTS is able to learn an approximate posterior distribution, i.e. a bootstrap distribution over the expected utility of the returns. DMCTS outperforms both C51 and Q-learning because a posterior over the expected utility of the returns is a sufficient statistic on which to base exploration. Moreover, learning a bootstrap distribution is an efficient yet compact approximation of a posterior distribution.

\subsection{Dangerous Deep Sea Treasure}
We now demonstrate that DMCTS can also learn successfully under the SER criterion as stated in Section \ref{sec:DMCTS}, even in an environment with stochastic transitions and rewards. For this, we adapt Deep Sea Treasure (DST), which is a commonly used benchmark for MORL algorithms under the SER optimisation criterion \cite{vamplew2011}. We introduce the Dangerous Deep Sea Treasure (DDST) environment, which is a stochastic variant of the DST problem. In DDST a submarine controlled by an agent searches for treasure on the sea bed where there are three objectives: treasure, damage and time. At certain states in the environment, the submarine can be attacked by a shark resulting in a negative reward in a separate objective, with probability $p_{shark}$. If the submarine receives a hit from a shark, the submarine becomes damaged. The submarine is destroyed if it accumulates a total of $-10$ damage, which terminates the episode. 

Equation \ref{eqn:unitvec_utilityfunc} describes the non-linear utility function we use to determine if DMCTS can learn a target Pareto optimal policy. In Equation \ref{eqn:unitvec_utilityfunc}, \textbf{r} is a reward vector, $\textbf{e} = \frac{r_{\dagger}}{|r_{\dagger}|}$ where $\textbf{r}_{\dagger}$ is a specified target vector, and $c$ is a constant we aim to maximise. The utility of $\textbf{r}$ is the maximum $c$ value where $\textbf{r} - c\textbf{e}$ is greater than $0$ for all objectives. The target vector, $\textbf{r}_\dagger$, is initialised to the desired vector we want to recover. 
\begin{equation}
    u(\textbf{r}) = \arg \max c : \textbf{r} - c\textbf{e} > 0 
    \label{eqn:unitvec_utilityfunc}
\end{equation}

\begin{figure}[t]
    \centering
    \includegraphics[width=0.35\textwidth]{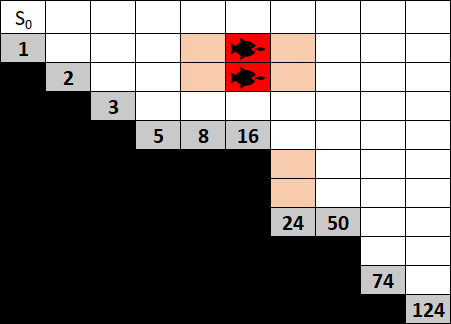}
    \caption{In DDST, states marked in red are terminal as the submarine is destroyed by a shark. The state in light red are non-terminal states where the agent has a probability, $p_{shark}$, of been hit by a shark. A hit by a shark causes -10 damage and the submarine is destroyed.}
    \label{fig:DDST}
\end{figure}

For this demonstration, we implemented scalarised Q-learning \cite{vanmoffaert2013scalarised} as that it is one of the most widely used SER algorithms \cite{vamplew2011}. DMCTS uses an extra exploration strategy which was outlined by Osband et al.\ \cite{osb2015bootstrapped}. For DMCTS, artificially generated returns of the environment are randomly sampled during the learning phase to ensure sufficient exploration. Q-learning has the following learning parameters: $\alpha = 0.1$, $\gamma = 1$ and a decaying $\epsilon = 0.998^{e}$, where $e$ is equal to the episode number \cite{vamplew2011, Mannion2017}. In the utility function we set $\textbf{r}_{\dagger}$ to $[54, 0, -14]$, where the objectives are ordered as follows: [treasure, danger, time]. We set $n_{exec} = 10$ and $p_{shark} = 0.5$. We run $10,000$ episodes per experiment.
\begin{figure}
    \centering
    \input{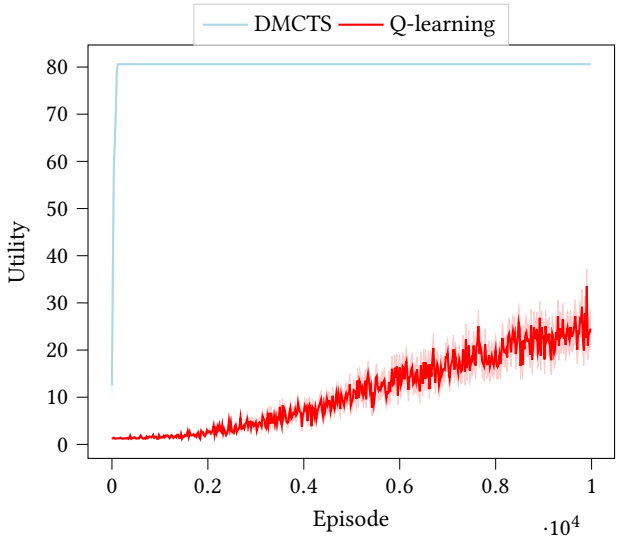}
    \caption{Results from the Dangerous Deep Sea Treasure environment using a non-linear utility function. DMCTS learns optimal utility for the specified target vector, $\textbf{r}_\dagger$.}
    \label{fig:ddst_utility}
\end{figure}


DMCTS converges to the optimal utility after $100$ episodes (Figure \ref{fig:ddst_utility}). This stands in contrast to scalarised Q-learning, which does not reach the optimal utility. 
Learning a bootstrap distribution over the expected returns under SER provides the agent with sufficient information to avoid states that represent negative utility (in this case, danger states), in order to maximise the known utility function. 
\section{Related Work}
Many risk-aware RL approaches seek to learn policies to maximise the expected return. Some research in this area focuses on learning policies which maximise the expected exponential utility \cite{Moldovan2012}. Other approaches take the weighted sum of the return and risk into consideration when learning policies \cite{Gosavi2009, Geibel2005}. Although most risk-aware RL approaches aim to maximise the expected utility, they often do not take into consideration the utility of the return of a full episode. It is also important to note that little research exists where decisions are made based on a learned distribution over the expected returns \cite{Morimura2010a, Morimura2010b} for risk-aware RL. 

As previously highlighted, the majority of RL research focuses on the SER criterion. Multi-objective MCTS (MOMCTS) \cite{Sebag2012} was shown to be able to learn a coverage set under SER. However, MOMCTS can only learn a coverage set in deterministic environments. Convex Hull MCTS \cite{Painter2020} is able to learn the convex hull of the Pareto front but focuses solely on linear utility functions. A number of other multi-objective MCTS methods exist \cite{Perez2013, Jongmin2018, Perez2015}, but no method has previously been shown to learn the Pareto front for both deterministic and stochastic environments for any unknown utility function.
An interesting opportunity for future work is the possibility of building on the methods of Wang and Sebag \cite{Sebag2012} and Painter et al.\ \cite{Painter2020} to extend DMCTS to learn the optimal coverage set under both SER and ESR for any unknown utility function.

A key argument in this paper is the expected utility of the future returns under ESR must be replaced with a posterior distribution over the expected utility of the returns. Bai et al.\ \cite{Bai2014} extend MCTS to maintain a distribution at each node using Thompson Sampling as an exploitation strategy. However, the work presented in this paper is significantly different. In their work, Bai et al.\ \cite{Bai2014} do not learn a posterior distribution over the expected utility of the return, apply their work to multi-objective settings, or incorporate the accrued returns as part of their algorithm. It is also important to note the C51 algorithm proposed by Bellemare et al.\ \cite{bellemare2017distributional} achieves state-of-the-art performance in single-objective settings and learns a distribution over the future returns. Abdolmaleki et al.\ \cite{Abdolmaleki2020} learn a distribution over actions based on constraints set per objective. This approach ignores the utility-based approach \cite{roijers2013survey} and uses constraints set by the user to learn a coverage set of policies where the value of constraints is dependent on the scale of the objectives. Abdolmaleki et al.\ claim setting the constraints for this algorithm is a more intuitive approach when compared to setting weights for a linear utility function. We theorise that if the user's utility function is non-linear, this approach would fail to learn a coverage set.

\section{Conclusion \& Future Work}
In this paper we propose a novel Distributional Monte Carlo Tree Search algorithm. DMCTS is able to learn optimal policies in MORL settings under both ESR and SER for both linear and non-linear utility functions in problem domains with stochastic rewards. DMCTS replaces the expected utility of the future returns with a bootstrap distribution over the utility of the returns, and achieves state-of-the-art performance in MORL domains under ESR. 
We achieve this by using a bootstrap distribution as an approximate posterior over the expected utility of the returns of the episode. It is our hope that this paper will inspire further work on algorithms that replace the expected returns with a distribution over the expected utility of the returns for risk-aware and ESR settings.

In this paper, the utility function is known at the time of learning or planning. In different MORL scenarios, the utility function can be unknown at the time of learning or planning \cite{roijers2013survey, radulescu2020survey}. In these scenarios, an algorithm must recover a coverage set of optimal policies. Multi-objective MCTS \cite{Sebag2012} can learn a coverage set for deterministic environments under SER. In future work, we aim to extend our DMCTS algorithm to be able to learn coverage sets for unknown utility functions under ESR and SER for stochastic environments. However, a coverage set of optimal policies under ESR has yet to be defined. We therefore hope to define the coverage set of optimal policies for ESR and extend DMCTS to learn it. 


\bibliographystyle{ACM-Reference-Format}  
\bibliography{references}  

\end{document}